\documentclass[5p,times]{elsarticle}
\usepackage[utf8]{inputenc}
\usepackage{algorithm}
\usepackage[noend]{algpseudocode}
\usepackage{amssymb}
\usepackage{bbold}
\usepackage{graphicx}
\usepackage{multirow}
\usepackage{makecell}

\graphicspath{ {./images/} }

\algnewcommand{\algorithmicand}{\textbf{ and }}
\algnewcommand{\algorithmicor}{\textbf{ or }}
\algnewcommand{\OR}{\algorithmicor}
\algnewcommand{\AND}{\algorithmicand}
\algnewcommand{\var}{\texttt}
\algnewcommand\algorithmicforeach{\textbf{for each}}
\algdef{S}[FOR]{ForEach}[1]{\algorithmicforeach\ #1\ \algorithmicdo}

\newcommand{\NOT}{\textbf{not }}

\usepackage{xcolor}

\makeatletter
\def\algbackskip{\hskip-\ALG@thistlm}
\makeatother

\begin{document}

\title{Scalable Data Point Valuation in Decentralized Learning}

\author[add1,add2]{Konstantin D. Pandl}
\author[add2]{Chun-Yin Huang}
\author[add2]{Ivan Beschastnikh}
\author[add2]{Xiaoxiao Li}
\author[add1]{Scott Thiebes}
\author[add1]{Ali Sunyaev}

\address[add1]{Karlsruhe Institute of Technology, Karlsruhe, Germany}
\address[add2]{University of British Columbia, Vancouver, Canada}

\date{May 2023}

\begin{abstract}
Existing research on data valuation in federated and swarm learning focuses on valuing \emph{client} contributions and works best when data across clients is independent and identically distributed (IID). In practice, data is rarely distributed IID. 

We develop an approach called DDVal for decentralized data valuation, capable of valuing \emph{individual data points} in federated and swarm learning. DDVal is based on sharing deep features and approximating Shapley values through a k-nearest neighbor approximation method. This allows for novel applications, for example, to simultaneously reward institutions and individuals for providing data to a decentralized machine learning task.

The valuation of data points through DDVal allows to also draw hierarchical conclusions on the contribution of institutions, and we empirically show that the accuracy of DDVal in estimating institutional contributions is higher than existing Shapley value approximation methods for federated learning. Specifically, it reaches a cosine similarity in approximating Shapley values of 99.969 \% in both, IID and non-IID data distributions across institutions, compared with 99.301 \% and 97.250 \% for the best state of the art methods. 

DDVal scales with the number of data points instead of the number of clients, and has a loglinear complexity. This scales more favorably than existing approaches with an exponential complexity. We show that DDVal is especially efficient in data distribution scenarios with many clients that have few data points - for example, more than 16 clients with 8,000 data points each. By integrating DDVal into a decentralized system, we show that it is not only suitable for centralized federated learning, but also decentralized swarm learning, which aligns well with the research on emerging internet technologies such as web3 to reward users for providing data to algorithms.

\end{abstract}

\maketitle

\section{Introduction}
Machine learning (ML) training requires large amounts of diverse and well-labelled data. This can be challenging for certain industries, like healthcare, where data movement is governed by privacy regulations~\cite{zou2019primer} and labelling the data requires experts who may be difficult to find and expensive~\cite{radsch2021your}. Generally, patients may own their data in certain jurisdictions~\cite{ballantyne2020should}, and using their data for training models can involve a lengthy consent process. Furthermore, there are few incentives for patients to share their data for AI systems, which can result in a lack of data~\cite{pandl2022reward}.

These challenges have sparked interest among practitioners and researchers in systems giving individuals control over their data and rewarding them for providing data to AI algorithms~\cite{hynes2018demonstration}. One research direction is to use blockchain-based token economies and web3 technologies~\cite{sunyaev2021token, jung2021mechanism}. Early approaches use a blockchain network in combination with trusted execution environments to process data in protected hardware enclaves~\cite{hynes2018demonstration}. This processing includes the training of ML models, the valuation of data points based on their influence on the ML task, and the inference of ML models. Thereby, model consumers requesting an ML model inference pay model providers and data providers to use ML models. The stream of research falls into a larger recent trend in practice that entities request compensation for providing their data to ML companies \cite{diaz_2023}. 

Another research direction tackles the problem of training data availability with federated learning (FL)~\cite{mcmahan2017communication}. FL enables institutions to collaboratively train ML models without sharing data. An FL process is either orchestrated through a central server, or increasingly in a peer to peer communication between clients. This is also referred to as swarm learning (SL), often implemented through smart contracts on a private blockchain network. SL improves on robustness and auditability of the FL process. First whitepapers by practitioners demonstrate the need to develop reward systems for institutions in blockchain-based SL environments~\cite{hpe}.

Extant research on incentivizing data contributions from individuals while preserving privacy is limited. One paper presents a concept based on trusted execution environments (TEEs), where data is pooled, valued, and ML models are trained~\cite{hynes2018demonstration}. However, it doesn't describe the technical system in detail, and TEEs have limited scalability and have the risk of releasing private information through side channel attacks. Generally, this raises the question if there are alternative ways to enable both, incentivizing data contributions from individuals while preserving their privacy. To preserve the privacy, we build upon decentralized learning techniques, while aiming to value data. Thereby, we differ from extant research~\cite{song2019profit, kumar2022towards} by focusing on valuing data points instead of individual contributions.

Recent research has developed reward systems for institutions in FL based on the quality of their contributions~\cite{pandl2022reward, witt2022decentral}, which is often computationally expensive to quantify. However, these reward systems are based on data valuation techniques that work best for independent and identically distributed (IID) data distributions~\cite{song2019profit, kumar2022towards}, which is unrealistic in practice. Furthermore, data at FL clients may originate from individuals - for example, in the case of hospitals participating in FL, the data may be provided by patients. It remains unclear how such reward systems can be designed to reward individuals corresponding to clients in FL or SL scenarios with resource limitations (computation, storage, network bandwidth), and how to best use potential benefits of blockchain and distributed ledger technology (DLT) (e.g., commitments of institutions through the staking of tokens)~\cite{pandl2020convergence}.

Such knowledge is important to design data provision incentives for industries such as healthcare and to unlock the full potential of AI. First research shows the utility and necessity of such systems, but does not provide a detailed technical architecture and relies on the use of special techniques such as trusted execution environments~\cite{hynes2018demonstration}. We aim to develop an approach for hierarchical data valuation named DDVal that can value data points in decentralized learning, instead of valuing contributions of institutions. This allows us to compute hierarchical rewards and data values of groups, such as contributions of patients with multiple data points or entire institutions. We integrate DDVal into a system where we translate the valued data into rewards, and develop an SL and token economy architecture in which we incorporate DDVal. Our core contributions are four-fold:

\begin{itemize}
  \item We develop the concept of data point valuation in decentralized learning and present a method named DDVal for it that we evaluate.
  \item We develop the concept of hierarchical data valuation in FL and show that we can also value contributions of institutions, which we compare with existing methods.
  \item We develop a smart contract system that integrates SL with a token economy that automatically incentivizes institutions for participating in decentralized learning in a transparent manner.
  \item We outline how the system can also help to incentivize patients for contributing data and enable trustworthy AI in healthcare.
\end{itemize}

The remainder of this paper is organized as follows. In Section~\ref{sec:background}, we provide a background into FL and SL, contribution valuation in FL, Shapley value estimation in centralized ML, and token economies. Then, we describe our system design in Section~\ref{sec:systemdesign}. Subsequently, we describe the method in Section~\ref{sec:experimentalmethod} to evaluate the system. We present the results in Section~\ref{sec:experimentalresults}. We provide a discussion in Section~\ref{sec:discussion}. Finally, we conclude in Section~\ref{sec:conclusion}.

\section{Background}\label{sec:background}
\subsection{Federated and swarm learning}
FL enables the training of ML models across separated data silos held by different clients. In its initial use case, it was proposed for training language models for smartphone keyboards~\cite{mcmahan2017communication,hard2018federated} while preserving the privacy of users.
An FL process consists of iterated communication rounds, each round comprising of four steps. First, the distribution of the global ML model to the clients. Second, the local optimization at each client's site. Third, the distribution of the gradients from the clients to the server. Fourth, the aggregation of the model and the gradients into an updated model, typically through averaging. Communication rounds are repeated until an ending condition is reached (e.g., no more improvements in predictive performance based on a test set). Since the actual training data does not leave the client's sites, FL can help to preserve the privacy of data owners. Therefore, the use of FL is nowadays widely debated in industries like healthcare~\cite{xu2021federated}.

SL is a further decentralized variant of FL. Instead of each client connecting to a centralized server, clients are connected to each other in a peer-to-peer fashion~\cite{warnat2021swarm}. SL is often implemented through a private blockchain network and a separated file sharing network. A blockchain is a network of replicated state machines enabling to store data in a decentralized network in a tamper-proof manner~\cite{kannengiesser2020trade}. In SL, ML models and gradients, or hashes thereof are stored on the private blockchain network. As such, it is also referred to in literature as blockchain-based FL. This helps to keep the record of model and gradient updates, and controls the SL process. Advantages of SL over FL are better auditability and explainability, higher network robustness, and access control through the blockchain network. As SL mainly affects the communication layer of the learning process, the models obtained from SL generally does not differ in its parameters or properties from FL. The decentralization can add complexity to the network, but is beneficial in certain industries where trustworthy ML is of high importance, like healthcare~\cite{warnat2021swarm, fan2021fairness, pandl2020convergence}. As complex ML models are typically too large to store on a blockchain, a common solution is to combine the blockchain network with an InterPlanetary File System (IPFS) network~\cite{zhao2020privacy}. IPFS is a peer to peer filesharing network where files are indexed with their hash. Thus, clients can retrieve files from the network if they know the hash of the file~\cite{benet2014ipfs}. For blockchain-based FL, the hashes of models and gradients are stored on-chain. Thus, all clients in the network receive the hashes and can retrieve the files through IPFS~\cite{pandl2020convergence}.

\subsection{Contribution valuation in federated learning}
Data valuation in FL is typically focused on valuing contributions of the individual clients. A dominant concept for data valuation is the Shapley value which is defined by the following equation, where $D$ is the set of the individual clients $z$ of size $N$, and $U$ is a utility function (e.g., area under the receiver operating characteristic, AUROC, of the ML model):

\[
\varphi_{Shap}(z_i) = \frac{1}{N} \sum_{S \subseteq D \setminus z_i }^{} \frac{1}{{N-1\choose |S|}} (U(S \cup z_i) - U(S))
\]

Following this definition, the SV uniquely satisfies three desirable properties which are highly relevant in the context of data valuation~\cite{jia2019efficient, ghorbani2019data}. First, group rationality, meaning the entire value gain of the FL consortium is completely distributed among all participating institutions: $U(D)=\sum_{z_i \in D}^{} \varphi(z_i)$.

Second, fairness, meaning the SV of an institution $z_i$ should be zero if the institution has zero marginal contribution to all subsets of the consortium: i.e., $\varphi(z_i) = 0$ if $U(S \cup z_i) = 0$, for all $S \subseteq D \setminus z_i$.
At the same time, the contribution quantification of two institutions $z_i$ and $z_j$ should have the same value if they both add the same performance to a subset of the consortium, meaning $\varphi(z_i) = \varphi(z_j)$ if $U(S \cup z_i) = U(S \cup z_j)$, for all $S \subseteq D \setminus z_i \setminus z_j$.

Third, additivity, meaning the Shapley values under multiple utilities sum up to the Shapley value under a utility that is the sum of all these utilities: $\varphi(v_1, z_i) + \varphi(v_2, z_i) = \varphi(v_1 + v_2, z_i)$.

In its canonical form, computing the SV requires the evaluation (training and testing) of every possible subset of entities, and requires $2^N-1$ evaluations for $N$ FL participants. Thus, computing the SV can be computationally highly expensive even for a small number of clients.

Therefore, researchers developed approximations to estimate SVs with less computational effort. Most of these approximation algorithms aim at efficiently approximating the trained model of coalitions~\cite{song2019profit}, some methods also aim at reducing the number of coalitions that need to be tested~\cite{liu2022gtg} or to speed up the testing process~\cite{kumar2022towards}. A first SV approximation method by Song et al.~\cite{song2019profit} works by training the largest coalition and summing up gradients to approximate smaller coalitions. Then, all approximated coalitions are tested. This massively reduces the training complexity from training $2^N-1$ coalitions to $1$ coalition. As such, the method is referred to as One-Round (OR) approximation. While still $2^N-1$ models need to be tested, the testing typically requires much less computational and communication effort than the training.

Another method by Kumar et al. named SaFE similarly in a first step works by training the largest FL coalition normally~\cite{kumar2022towards}. Afterward, each client extracts per-datum feature vectors and trains a logistic regression (LR) model classifier based on these. Then, $2^N-1$ coalitions of combinations of the LR classifier are combined and tested. Testing such an LR classifier on the feature vectors of the test dataset is a very fast process, as such, the authors show that their approach scales up to 20 clients and beyond~\cite{kumar2022towards}.

The accuracy of these SV approximation method is typically measured by the cosine similarity between the vector of the canonical SVs, and the approximated SVs~\cite{kumar2022towards, song2019profit}. Cosine similarity effectively describes the cosine of the angle of the two vectors, where a value of 1 is achieved for an exact approximation, and a value of -1 would describe completely opposite approximation values. The following equation describes the formula to compute the cosine similarity of two vectors $\mathbf{a}$ and $\mathbf{b}$.

\[
S_C(\mathbf{a},\mathbf{b}) = \frac{\mathbf{a} \cdot \mathbf{b}}{|\mathbf{a}||\mathbf{b}|}
\]

\subsection{Shapley value estimation in centralized machine learning}\label{sec:svcml}
In centralized ML, SV estimation is typically used to value individual data points during the training process. Contrary to FL, the size of the coalition is much larger (i.e., tens of thousands, hundreds of thousands or even more data points). As such, computing the SV in its canonical form is out of reach and efficient approximations are required. One popular approximation method for centralized ML is the k-nearest neighbor (KNN) SV approximation method~\cite{jia2019efficient}. For this, the neural network (NN) is trained with the entire set of training data points in a first step. Then, for the training and test points, deep features are extracted. Deep features are representations of the training and test data points by extracting the last layer activations of the NN, similar to the approach for FL by Kumar et al. Afterward, a KNN classifier is obtained that replaces the NN. Thus, only a small set of training data points influence decisions on local inference points, instead of the entire training dataset. As a result, not all subsets have to be considered, which speeds up the computation~\cite{jia2019efficient}. The SVs $\varphi$ can then be computed recursively for the $N$ data points using a test data point, and a KNN-parameter $K$ along with the following formulas:

\[
\varphi (z_N) = \frac{\mathbb{1}[y_{z_{N}} = y_{test}]}{N}
\]

\[
\varphi (z_i) = \varphi (z_{i+1}) + \frac{\mathbb{1}[y_{z_{i}} = y_{test}] - \mathbb{1}[y_{z_{i+1}} = y_{test}]}{K} \frac{min(K,i)}{i}
\]

We also illustrate the procedure in Algorithm~\ref{alg:knn-sv}. In practice, there are multiple test data points, and the resulting SV of a training data point is averaged across the test points, following the additivity property of the SV. Generally, the complexity of computing the KNN SV is $O(N \log N)$ for $N$ data points to value.
Successful applications include medical imaging in healthcare, for example, to efficiently identify mislabelled images, or to remove unnecessary data and better protect patients' privacy~\cite{tang2021data,pandl2021trustworthy}.

\begin{algorithm}
    \caption{ComputeSVs function (KNN-SV~\cite{jia2019efficient})}\label{alg:knn-sv}
    \textbf{Input}: training data:$D = \{(x_i,y_i)\}_{i=1}^{N}$, test data: $D_{test}=\{(x_{test,i},y_{test,i})\}_{i=1}^{N_{test}}$\\
    \textbf{output}: The SV $\{\varphi_i\}_{i=1}^N$

    \begin{algorithmic}[1]

    \For{$j\gets1$ to $N_{test}$}
    \State $(\alpha_1, \cdots, \alpha_N) \gets$Indices of training data in an ascending order using $d(\cdot,x_{test})$;
    \State $\varphi_{j,\alpha_N} \gets \frac{\mathbb{1}[y_{\alpha_N}=y_{test}]}{N}$;
     \For{$i\gets N-1$ to $1$}
     \State $\varphi_{j,\alpha_i} \gets \varphi_{j,\alpha_{i+1}} + \frac{\mathbb{1}[y_{\alpha_i}=y_{test,j}] - \mathbb{1}[y_{\alpha_{i+1}}=y_{test,j}]}{K} \frac{min\{K,i\}}{i}$;
     \EndFor
    \EndFor
    
    \For{$i\gets1$ to $N$}
    \State $\varphi_i \gets \frac{1}{N_{test}} \sum_{j=1}^{N_{test}} \varphi_{j,i}$;
    \EndFor
    
    \end{algorithmic}
\end{algorithm}

\subsection{Token economies}
Token economies enable the transfer of ownership of assets (e.g., cryptocurrencies) between agents (e.g., individuals, organizations) based on distributed ledger technology (DLT, i.e. blockchain)~\cite{sunyaev2021token}. Thereby, token economies reduce the need for trusted third parties and enable novel business models and transparent business processes, including monetization of data for individuals. This can support collaboration and cooperation between agents. In many cases, token economy systems comprise of multiple distributed ledgers, each optimized for specific use cases, and cross-chain technology for communication between distributed ledgers~\cite{sunyaev2021token}.
An essential part of token economies are the tokens which are used for the representation of assets, such as monetary value. Traditional cryptocurrencies like Bitcoin or Ethereum are highly volatile and may be impractical for payment activities. Therefore, blockchains like Ethereum host stablecoins, which are pegged to a stable currency (e.g., the US-Dollar) and are aimed to have little to no volatility in their price~\cite{moin2020sok}. Such stablecoins on Ethereum are often implemented with the ERC-20 standard. This provides a standardized interface and libraries to develop smart contracts interfacing ERC-20 tokens.

Early research proposes token economies to incentivize clients to contribute in FL~\cite{pandey2022fedtoken, liu2020fedcoin}. For example, in FedCoin~\cite{liu2020fedcoin}, the authors include SV computation in FL in the block generation of a blockchain, and pay off FL clients based on the quality of the contributions in their clients. Further early research proposes the use of blockchains based on trusted execution environments to have a secure computing area for data valuation~\cite{hynes2018demonstration}. Based on the data values, model consumers can then pay model and data providers through a blockchain system.

\section{System design}\label{sec:systemdesign}
\subsection{System overview}
Our core contribution is the development of a novel approach for data valuation in FL and SL that can value individual data points named DDVal. We also demonstrate the integration of DDVal into SL and a token economy through a novel system. This consists of ML aspects, and network (blockchain and IPFS) aspects.
The core of the ML aspect is the training of the federated coalition, and an approach to hierarchically value data based on the KNN-SV approach. We describe it in detail in Section~\ref{chapter:hdva}.

For SL, we use a private blockchain network with a smart contract and a private IPFS network described in Section~\ref{chapter:sla}. For the token economy, we design and deploy a smart contract for a public blockchain, described in Section~\ref{chapter:tea}. The entire process of the SL and token economy is also illustrated in the sequence diagram in Figure~\ref{fig:sequence} in Section~\ref{sec:integration}.

\subsection{Hierarchical data valuation approach}\label{chapter:hdva}
First, we aim to value data points for FL and SL. A naive approach would be to value client contributions first and then have each client value their data points as in centralized ML, and multiply both values. However, this does not capture the utility of collaboration of multiple clients in FL. Therefore, we take a more sophisticated approach, inspired by the KNN-SV method for centralized ML. Based on the individual SVs for the data points, we also aim to derive contributions of groups of data points (e.g., clients).

In the first step, the FL or SL process is conducted with the entire set of clients in repeated communication rounds, until an ending condition is reached, which can be defined as a plateau of an evaluation metric (i.e., AUROC) on a validation dataset is reached. Thus, all clients obtain the same global model.
Afterward, the special procedure for the hierarchical data valuation begins, which is described in Algorithm~\ref{alg:fl-server} from the FL server perspective and Algorithm~\ref{alg:fl-client} from the client perspective. To obtain the deep features in the first place, every FL client first runs an inference of each of their training and test data points using the global model. They store the activations of the last output layer of the neural network, referred to deep features of the data points, during this inference process. Then, the clients submit the deep features to the server in the centralized FL case or to the other clients in a decentralized SL case. Then, we compute the SV for every training data point based on the procedure for the KNN-SV described in Section~\ref{sec:svcml}. Furthermore, we normalize the SVs such that the sum matches the utility of the largest coalition, as described by the following formula:

\[
\varphi_{AUC} (z_i) = \varphi (z_i) \frac{AUC_{coalition}-0.5}{\sum_{i} \varphi (z_i)}
\]

Thereby, we define the effective utility of the coalition as their AUROC score minus 0.5, since a random classifier has an AUROC score of 0.5, and can range from 0 to 0.5.

In the case of a centralized, server-based FL, the server sends the SVs of the specific data points to the corresponding clients. In the decentralized SL case, the clients can compute the SVs themselves. To estimate contributions for arbitrary hierarchies (i.e., groups of data points), we summarize the SVs of the individual data points: $\varphi_{AUC}(D) = \sum_{z \in D} \varphi_{AUC}(z)$ for a group $D$ of data points. The intuition behind this is that rewards are typically distributed proportionally to the SVs, and the sum of the rewards should be proportional to the group contributions~\cite{jia2019efficient}. In practice, for example, this means contributions by a group of patients or from an entire hospital can be valued. For details, we refer to Algorithm~\ref{alg:fl-server} from the server perspective and Algorithm~\ref{alg:fl-client} from the client perspective. Furthermore, we refer to Algorithm~\ref{alg:knn-sv} in the background section on the procedures to compute the SVs.

DDVal also ensures that the same data points at different clients get the same SV. This is because the data valuation is computed centrally for all deep features, regardless of what clients they originate from. For the same deep features, the algorithm will yield in the same SV, satisfying the fairness condition of the SV.

\begin{algorithm}
    \caption{Data valuation procedure for FL from the view of the server.}\label{alg:fl-server}
    \textbf{Notation:} \\
    $\Psi = \{\psi_0, \cdots, \psi_N\}$: Global model for each round\\
    $AUCs$: Collection of validation results\\ $K$: k parameter in the KNN method
    
    \begin{algorithmic}[1]
    \While {\NOT endingCondition}

    \For {each client $c$}
    \State $\nabla_c \gets trainLocally(\psi)$
    \EndFor
    \State $\psi_{i+1} \gets aggragate(\nabla, \psi_i)$
    \For {each client}
    \State $AUCs \gets validate(\psi_{i+1})$
    \EndFor
    \If{$min(AUCs_{i-endingConditionPeriod:i}) < AUCs_i$}
    \State $endingCondition \gets 1$
    \EndIf
    \EndWhile
    
    \For {each client $c$}
    \State $F_{tr}, y_{tr}, F_{te}, y_{te}  \gets client.extractDeepFeatures(\psi_{i+1})$
    \EndFor
    \State $SVs \gets computeSVs(K, F_{tr}, y_{tr}, F_{te}, y_{te})$

    \end{algorithmic}
\end{algorithm}

\begin{algorithm}
    \caption{Data valuation procedure for FL from the view of the client.}\label{alg:fl-client}
    \textbf{Notation:} \\
    $(X_{tr}, y_{tr})$: Training data pairs \hspace{.5em} $(X_{te}, y_{tr})$: Testing data pairs\\ 
    $F_{tr}$: Deep training features \hspace{1em} $F_{te}$: Deep testing features\\
    $\psi$: Global model \hspace{5em} $\nabla$: Gradients
    
    \begin{algorithmic}[1]

    \Function{trainLocally}{$\psi$}
    \hspace*{\algorithmicindent}\State $\nabla$ = train($X_{tr}$, $y_{tr}$, $\psi$)
    \State \textbf{return} $\nabla$
    \EndFunction \\
    
    \Function{validate}{$\psi$}
    \hspace*{\algorithmicindent}\State $AUC$ = validate($X_{te}$, $y_{te}$, $\psi$)
    \State \textbf{return} $AUC$
    \EndFunction \\
    
    \Function{extractDeepFeatures}{$\psi$}
    \State $F_{tr}$ = $\psi(X_{tr})$
    \State $F_{te}$ = $\psi(X_{te})$
    \State \Return $F_{tr}$, $y_{tr}$, $F_{te}$, $y_{te}$
    \EndFunction
    
    \end{algorithmic}
\end{algorithm}

\subsection{Swarm learning approach}\label{chapter:sla}
For the SL, we set up a private Ethereum network based on a proof of authority consensus mechanism. This blockchain fulfills the goal of having a peer-to-peer connection between clients to communicate, whereby the clients are not necessarily fully connected. Furthermore, the private blockchain enables control of who can access the network and serve as an immutable ledger for storing transactions. We use Go-Ethereum (Geth) version 1.10.25 in our experiments.

For this private blockchain, we developed a smart contract that orchestrates the SL process and the subsequent data valuation using the Solidity programming language. The code is included in our GitHub repository\footnote{https://github.com/kpandl/Scalable-Data-Point-Valuation-in-Decentralized-Learning}. Furthermore, we set up a private IPFS network that enables to transfer larger files (i.e., model gradients in the orders of several Megabytes, and deep feature vectors) between the clients, as blockchain networks like Ethereum lack the capability to store such large files in transactions. Through IPFS, clients can publish a file in the network, and other clients can request the transfer of the file from the client through the hash. Such a combined setup of a private Ethereum network and a private IPFS network has proven to be successful for SL tasks~\cite{benet2014ipfs, pandl2020convergence}. For the SL smart contract, the owner (i.e., the institution submitting the contract to the blockchain) specifies allowed Ethereum addresses that are whitelisted to participate in the SL process, and specifies a hash for the initial ML model. Once created, the whitelisted institutions can signal that they are ready to start the SL process in the contract. Then, through the smart contract, the institutions iteratively exchange model gradient hashes and model test results in communication rounds. The actual model and model gradients are exchanged through the IPFS network. The smart contract then ends the SL process once the ending condition is reached, in our implementation no improvement in the test result over the last 10 communication rounds.

\subsection{Token economy approach}\label{chapter:tea}
We aim to extend the SL approach through a token economy that incentivizes institutions and also individuals (e.g., patients) corresponding to institutions for providing their data for the SL process. For transferring assets of monetary value, we generally want to use established, large and trustworthy networks. Large, public permissionless blockchains, such as the public Ethereum network, offer a high transaction integrity and an active ecosystem of tokens representing monetary values. For research purposes, we use the public Ethereum Sepolia testnet. The public Ethereum networks host a large ecosystem of decentralized finance applications running on smart contracts, including stablecoins which are pegged to the value of a non-cryptocurrency such as the US dollar and, thus, typically have a low volatility in their value, especially when compared to traditional cryptocurrencies. This is desirable in smart contract applications in industries such as healthcare.

Besides the smart contract on the private blockchain, we also submit a smart contract to the public blockchain. Initially, an institution submits the compiled smart contract to the blockchain, and funds the smart contract with an ERC-20 stablecoin. Afterward, the SL is conducted on the private networks. We extend the processes on the private blockchain through DDVal, which is conducted subsequent to the SL. Afterward each institution conducts the data valuation, and the institutions know the SVs of other institutions - each institution should receive the same results of the computation. The institutions then submit these SVs to the smart contract on the public blockchain, and this smart contracts conducts the payment of the rewards to institutions. Any leftover tokens (i.e., if the result of the SL is not perfect in terms of achieving an AUROC of 1) will be returned to the depositors - thus, it is likely that the depositors will also receive funds back.

Institutions can then pass the rewards entirely or partially further toward individual data providers, such as patients in a healthcare setting, in proportion to the sum of the SVs of their data points. Such payments can either be conducted through selling the tokens on an exchange and using the traditional banking system, or with relatively scalable blockchains with potentially lower transaction costs than Ethereum (e.g., Litecoin).

\section{Experimental method}\label{sec:experimentalmethod}
To evaluate DDVal, we answer four research questions that evaluate it from different angles, which we describe below. In our evaluation, we use a combination of three chest X-ray datasets of NIH~\cite{wang2017chestx}, CheXpert~\cite{irvin2019chexpert}, MIMIC-CXR~\cite{johnson2019mimic} for the ML tasks. These datasets are widely used in ML healthcare research and allow us to imitate realistic settings of non-IID data distributions. As for the ML model, we use a 121 layer DenseNet architecture~\cite{huang2017densely}, pre-trained on the ImageNet dataset. This architecture has shown to work well with CXR datasets~\cite{ke2021chextransfer}, and also with KNN-based data valuation in centralized ML~\cite{pandl2021trustworthy}. Thereby, we extract the deep features from the last, fourth dense block.

With our experimental method, we aim to tackle the following RQs:

\textbf{RQ1. How effective can DDVal value data points?}
An effective data valuation mechanism can successfully differentiate between data points beneficial for the ML task and data points harming the ML task by assigning values to the data points. The values can be in the form of SVs which have desirable properties for data valuation, as described above.

Computing canonical SVs is too computationally complex for data valuation tasks, where we have tens of thousands or even more data points~\cite{jia2019efficient, ghorbani2019data}. As such, we cannot compare the approximated SVs with the canonical SVs. However, we can focus on conducting two other experiments that allow us to draw conclusions about the quality of the SV estimations. In the first experiment, we randomly flip labels of data points. Specifically, we randomly flip between 0 \% and 25 \% of the labels of training data points depending on the client. Each data point has 8 different labels, so some clients have multiple flipped labels. Based on research on data valuation for centralized ML, we would expect data points with flipped labels to have lower SVs on average than data points without flipped labels~\cite{ghorbani2019data}. Second, we aim to analyze the SVs of the individual data points depending on their labels. Real-world datasets, like CXR datasets, often have multidimensional labels. For example, a CXR scan of a patient may show multiple medical conditions. Based on the knowledge from experiments in centralized ML, we would expect scans with medical conditions to have higher SVs than scans without~\cite{pandl2021trustworthy}.

\textbf{RQ2. How accurately can DDVal value contributions of institutions?}

Compared to number of data points, the order of the number of institutions in FL is typically much smaller (i.e., 2-100 institutions in cross-silo FL~\cite{kairouz2021advances}). As such, we can compute canonical SVs and compare our hierarchical contribution index with the canonical SVs. Furthermore, extant research proposes methods for SV estimation for clients in FL, which we can also compare. 

In our accuracy comparison experiments, we use 3 clients, meaning we need to evaluate 7 different coalitions to compute the canonical SV, which is computationally feasible. We repeat our experiment 12 times where the data distribution across clients depends on a different random seed. This allows us to get statistically significant results in the forms of a mean and confidence intervals. Thereby, we measure the SV estimation accuracy using the cosine similarity, a commonly used metric for this purpose~\cite{song2019profit,kumar2022towards}.

Since non-IID data distributions are challenging for successful FL implementations and data valuation in FL in particular, we analyze both, an IID and a non-IID case. FL and SL are especially debated in the healthcare industry, thus, we sample the training data from three different chest X-ray datasets: NIH~\cite{wang2017chestx}, CheXpert~\cite{irvin2019chexpert}, MIMIC-CXR~\cite{johnson2019mimic}. These datasets have varying input and label distributions. In the IID case, we have 3 clients with 27,999 scans consisting of 9333 scans sampled from each dataset. We make sure that a patient with potentially multiple scans is only associated to one client. In the non-IID case, each client again has 27,999 scans, but each client samples the scans from only one of the datasets.

Additionally, we analyze the institutional contributions in the same label-flip experiment from RQ1. Thereby, we expect the contributions of institutions to be smaller for clients with a higher percentage of flipped labels.

\textbf{RQ3. What is the scalability of DDVal?}

To analyze the scalability of different SV approximations, we conduct runtime experiments with different dataset sizes to value. Generally, the canonical SV computational effort is exponential to the number of entities (i.e., data points or FL participants). While approximations are more efficient, they generally also require higher computational effort with more entities. Thus, SV estimations of FL clients generally get slower with more clients.
As discussed, the data valuation approach in DDVal scales with regard to the number of data points, instead of the number of clients.

In our scenario, the complexity based on theory is $N \log N$ for $N$ data points~\cite{jia2019efficient}. As a result, it does not complexity-wise matter if these data points are distributed among few or many clients. As such, we expect DDVal to be particularly suited for decentralized learning cases with many clients and relatively few data points per client. We aim to validate the theoretical complexity with our experiments and compare it with existing SV-based client valuation approaches that scale with the number of clients~\cite{song2019profit,kumar2022towards}. To compare these approaches with DDVal, we assume a fixed number of 8,000 data points per client in the experiments. In the experiment, we consider the scalability with regard to the number of data points. For example, 16,000 data points refer to an experiment with 2 FL clients, and 32,000 points refer to an experiment with 4 FL clients. The test dataset size in the scalability experiments is 3,600 scans.

In the experiments, we measure the computational time an FL server needs to perform the data valuation, and we do not include the time required at the clients' site (e.g., deep feature extraction in SaFE and DDVal). The reasoning behind this is the fact that client-level computations are generally fast and a client only needs to perform computation for their own data, which is done in parallel across the clients. The complexity of the server computations, however, which values contributions from all clients, can grow quickly. In a decentralized SL scenario, the tasks of the server are replicated across (all) clients, thus the work of the FL server is conducted there as well.

\textbf{RQ4. How can we apply DDVal in a healthcare decentralized learning scenario to reward institutions and patients for providing data?}

This question focuses on distributed system aspects. As such, we run the experiments using a set of servers, instead of a high performance computing cluster. Thereby, we conduct the subsequent processes of SL, data valuation, and reward allocation and distribution.

SL describes the concept of peer to peer communication in decentralized learning, instead of a server-based communication. A common implementation of SL involves a private blockchain network~\cite{warnat2021swarm}. The servers run a private blockchain node and an IPFS node. They also connect to a public blockchain node. In the experiments, we aim to deploy the smart contracts and evaluate the runtime and monetary costs of the processes end-to-end.

Thereby, there are two types of user roles in the system. First, an orchestrator that publishes the smart contract on the public and private blockchain for the token economy and SL, and funds the smart contract on the public blockchain. Second, the institutions. An institution may also take the role of an orchestrator.

We develop and implement a system architecture integrating DDVal into SL and a token economy. Over the entire process, we measure the transaction cost for the orchestrator and the participating institutions. Furthermore, we measure the passed time. The goal for both measurements is to determine if the quantities are reasonable for the practical use. For the time, we can assume so if the overhead of the collaborative, decentralized learning is much shorter than the time required for the actual model training process. For the transaction cost, we can also compare the public blockchain costs to typical costs that large health institutions have.

\section{Experimental results}\label{sec:experimentalresults}

\subsection{Effectivity in valuing data points}\label{section:valDatPoint}

First, we focus on the label flip experiment, where we have 6 clients with 28,000 data points each, and each data point has a different number of flipped labels by chance. When looking at the individual data points and their SVs in Table~\ref{table:rq2}, data points with no flipped labels have the highest SVs with a mean of $2.43 \cdot 10^{-6}$. The more labels that are flipped for a data point, the smaller its SV is. The mean SV is negative for data points with 5 or more flipped labels. This is a first indication that the SV estimation provides reasonable results.

\begin{table}[h]
\centering
\caption{Mean SVs of individual data points of all clients in one experiment with flipped labels.}
\begin{tabular}{|c|c|}
\hline
{\bf Number of flipped labels} & {\bf SV $\cdot 10^{6}$} \\
\hline
0 & 2.43 ± 1.22 \\
1 & 1.88 ± 0.53 \\
2 & 1.34 ± 0.95 \\
3 & 0.74 ± 0.80 \\
4 & 0.14 ± 0.31 \\
5 & -1.17 ± 0.64 \\
6 & -2.17 ± 1.59 \\
\hline
\end{tabular}
\label{table:rq2}
\end{table}

Next, we focus on the mean SVs of data points depending on their labels, without any label flips. In the experimental setup, we have 3 clients with 28k data points each. As such, the SVs are not directly comparable to those in the previous experiment, and they are generally higher because the comparable utility of the trained global model is distributed among only half of the data points. The results are shown in Table~\ref{table:rq4}. Thereby, data points have the lowest mean SV of $1.1 \cdot 10^{-6}$ in cases when the label indicating a finding exists is set to true, but none of the 7 medical condition labels is set to true (i.e., an undefined medical condition is shown in the scan). The second-lowest mean SVs are for data points without any medical condition. Besides these two special cases, the SV is generally higher the more medical conditions that are present in the scan. Thus, the highest mean SV is for the scans with 4 medical conditions with $1.729 \cdot 10^{-5}$.

\begin{table}[h]
\centering
\caption{Mean SVs of data instances with multiple dimensions.}
\begin{tabular}{|c|c|c|}
\hline
\makecell{{\bf Finding existing }\\{\bf label }} & \makecell{{\bf Number of medical }\\ {\bf condition labels }} & {\bf SV } \\
\hline
False & 0 & $2.892 \cdot 10^{-6}$ \\
True & 0 & $1.100 \cdot 10^{-6}$ \\
True & 1 & $3.932 \cdot 10^{-6}$ \\
True & 2 & $8.117 \cdot 10^{-6}$ \\
True & 3 & $1.424 \cdot 10^{-5}$ \\
True & 4 & $1.729 \cdot 10^{-5}$ \\
\hline
\end{tabular}
\label{table:rq4}
\end{table}

\subsection{Valuing institutional contributions}

\begin{figure}[h]
\centering
\includegraphics[scale=0.6]{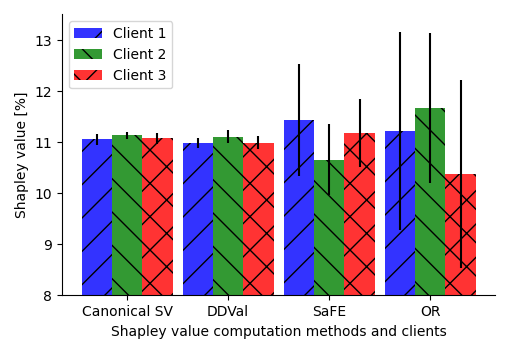}
\label{fig:rq1IID}
\caption{Mean SVs and 95\% confidence intervals of the clients, canonical and 3 approximations in the IID case.}
\end{figure}

The results of the institutional contribution valuation in the IID data distribution case are shown in Figure~\ref{fig:rq1IID}. There, the mean canonical SVs for the three institutions are relatively uniform and range from 11.047 \% to 11.125 \%. In DDVal, the SVs range from 10.984 \% to 11.105 \%, compared with 10.649 \% to 11.423 \% in SaFE and 10.371 \% to 11.213 \% in the OR approximation method. The 95 \% confidence interval range of the canonical SVs and DDVal is relatively small with 0.192 \% respectively 0.240 \% for DDVal, and it is 1.64 \% for SaFE and 3.503 \% for the OR approximation.

\begin{figure}[h]
\centering
\includegraphics[scale=0.6]{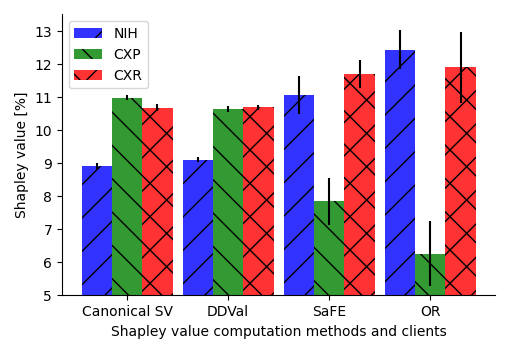}
\label{fig:rq1nonIID}
\caption{Mean SVs and 95\% confidence intervals of the clients, canonical and 3 approximations in the non-IID case.}
\end{figure}

The results of a similar experiment but conducted with a non-IID data distribution are shown in Figure~\ref{fig:rq1nonIID}. Across different clients using the same approximation method, the canonical SVs fluctuate more than in the IID case. They are the lowest for the NIH dataset with a mean of 8.918 \%, and the highest for the CheXpert dataset with a mean of 10.969 \%. In DDVal, NIH has the lowest SV with a mean of 9.101 \%, and MIMIC is slightly higher than CheXpert with 10.683 \% respectively 10.634 \%. In both other approximations, the CheXpert client has a relatively low SV of 7.831 \% (SaFE) and 6.246 \% (OR), whereas the approximated SVs of the NIH clients (7.831 \% and 6.246 \%) and MIMIC clients (11.682 \% and 11.890 \%) are higher than the canonical SVs and the estimations of DDVal.

To better compare the accuracy of the 3 approximations compared with the canonical SVs, we compute the cosine similarity based of these values. The cosine similarities of the approximations with the canonical SVs in both, the IID and non-IID cases, are shown in Table~\ref{table:rq1cosinesim}. Remarkably, DDVal has the highest mean cosine similarity of 99.969 \% and performs equally well for IID and non-IID data distribution. SaFE also has a relatively high cosine similarity with a mean of 99.301 \% in the IID case, but is negatively affected by a non-IID data distribution, where it drops to 97.250 \%. Similarly, for the OR approximation method, the cosine similarity drops from 97.189 \% to 93.753 \%.

\begin{table}[h]
\centering
\caption{Cosine similarities, means and 95 \% confidence intervals.}
\begin{tabular}{|c|c|c|}
\hline
\makecell{ {\bf SV approximation} \\ {\bf method} } & {\bf IID [\%] } & {\bf Non-IID [\%] } \\
\hline
DDVal & 99.969 ± 0.006 &  99.969 ± 0.010 \\
SaFE & 99.301 ± 0.403 & 97.250 ± 0.906 \\
OR approximation & 97.189 ± 0.756 & 93.753 ± 1.371 \\
\hline
\end{tabular}
\label{table:rq1cosinesim}
\end{table}

The reason why DDVal performs equally well for IID and non-IID data distributions is the fact that the deep features from all sites are collected and valued at one site. As such, the data partitioning influence diminishes. Contrary, creating approximated models in the OR method by accumulating gradients over different rounds generally performs inaccurately when clients have highly dissimilar gradients, as such, the cosine similarity drops in non-IID data distribution settings.

\begin{figure}[h]
\centering
\includegraphics[scale=0.6]{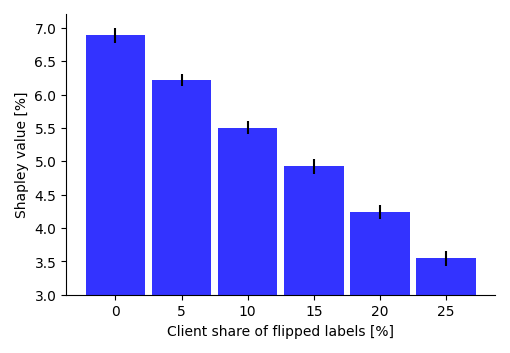}
\label{fig:rq2}
\caption{Label flip experiment, mean SV and 95\% confidence intervals by client.}
\end{figure}

As a last part in the institutional valuing section of DDVal, the results of the label flip experiment from Section~\ref{section:valDatPoint}, analyzed on an institutional hierarchical level, are shown in Figure~\ref{fig:rq2}. The client without any flipped labels has the highest mean SV with 6.884 \%. For the next 5 clients with an increasing share of 5 \% flipped labels, the mean SV constantly decreases to 3.548 \% for the client with a share for 25 \% flipped labels. For each 5 \% share increase, the SV decrease appears relatively constant, ranging from 0.58 to 0.713 percentage points in absolute terms. The confidence intervals are of relatively constant size in absolute range, ranging from 0.176 \% to 0.232 \%. Thereby, in relative terms to the mean SVs, the confidence intervals are larger for clients with smaller SVs (e.g., 6.37 \% for the client with 25 \% share of flipped labels vs. 3.14 \% for the client with no flipped labels).

\subsection{Scalability}
To measure the SV approximation compute time for client level methods, we assumed 8,000 data points per client. By definition, we need at least 2 clients in collaborative learning. Therefore, we measure the computational time for the client level methods SaFE and OR for 16,000 data points, and subsequently in 16,000 data point increments, which is equivalent to two more clients.

\begin{figure}[h]
\centering
\includegraphics[scale=0.6]{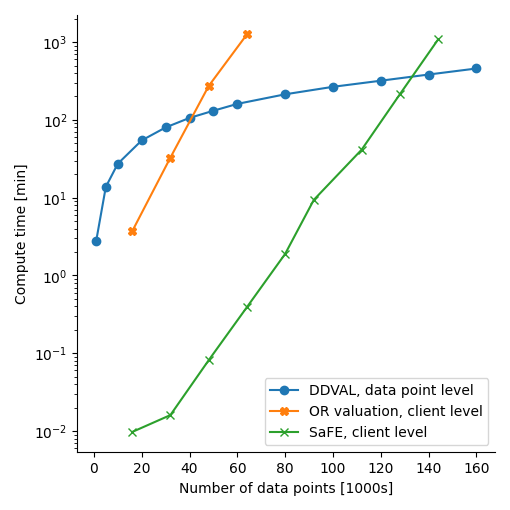}
\label{fig:rq3}
\caption{Scalability experiment. For the client level approximations, we assumed a fixed training dataset client size of 8,000 scans per client. The y-axis is logarithmic.}
\end{figure}

The results of the scalability experiment are shown in Figure~\ref{fig:rq3}. As the range of values on the y-axis is wide, we made it logarithmic. 

In the figure, the exponential complexities of the OR method and SaFE and the loglinear complexity of DDVal become appearant. For a range of up to ca. 5 clients with each client having 8,000 data points, the OR method is generally faster than DDVal valuing 40,000 data points. For more clients, DDVal is faster. We computed for the OR method up to 8 clients, which took ca. 21 hours to compute. By comparision, the SaFe method has similar exponential complexity but starts from a much lower baseline. As such, it took only ca. 23 seconds of compute time in the same scenario, whereas DDVal took ca. 2.5 hours.

The crossover in the required computational time for data valuation between DDVal and SaFE is at above 16 clients or 128,000 data points. In the most comprehensive experiment for SaFE, we valued contributions of 18 FL clients with a total amount of 144,000 data points in ca. 18 hours. For the most comprehensive experiment with DDVal, we had 20 clients with collectively 160,000 data points in less than 8 hours.

For 10 clients, the valuation time took 1.85 minutes. This compares to 272 minutes for DDVal. For the OR valuation, the computing took too long on our cluster, but it terminated after 1255 minutes for the setting with 8 clients. However, SaFE grows exponential in its complexity, compared with loglinear complexity of DDVal. In scenarios with more data points and clients, or fewer data points per client than 8,000, DDVal can be favorable over SaFE in terms of computational effort. For few clients and data points, the OR approximation is faster than DDVal. For 6 clients totaling 60,000 data points, the KNN-SV based valuation is faster with ca. 3.9 hours vs. 4.5 hours for the OR approximation. Based on theory, the OR approximation has exponential complexity like SaFE, requiring much more compute time for a rising number of clients.

\subsection{Integration into a swarm learning token economy}\label{sec:integration}

\begin{figure}[h]
\centering
\includegraphics[scale=.33]{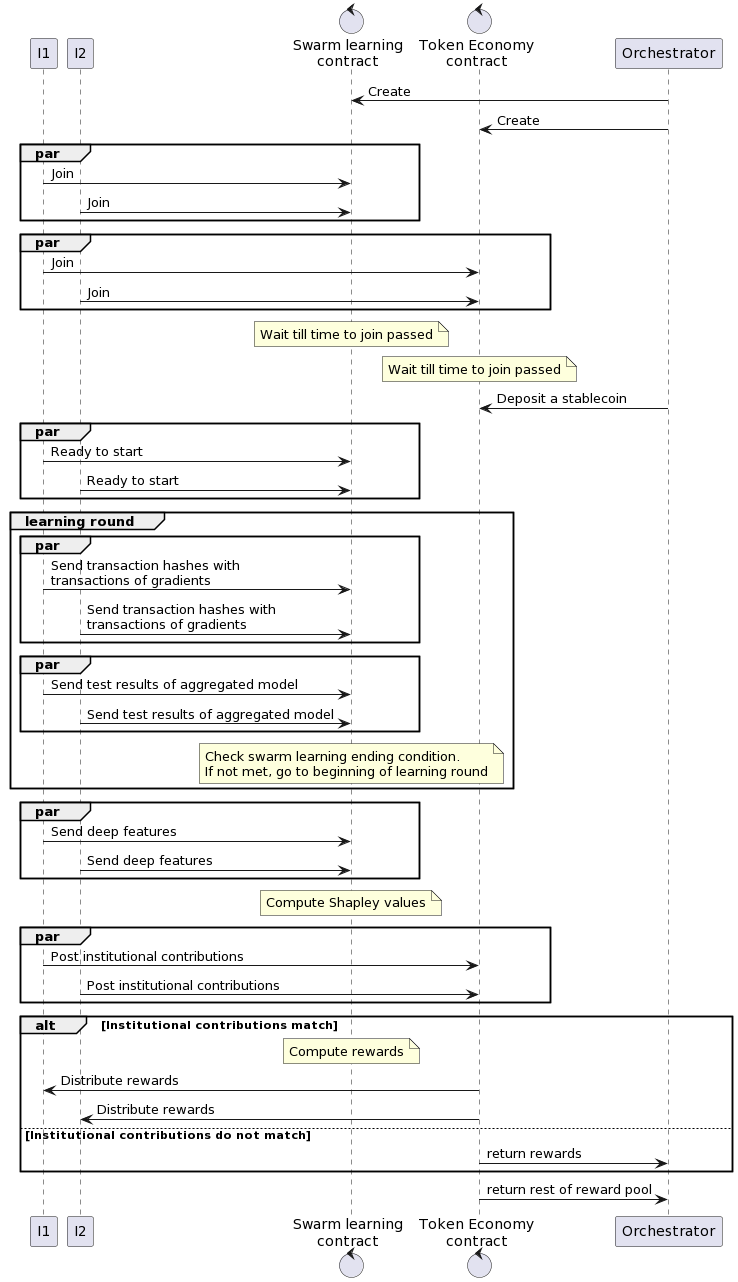}
\label{fig:sequence}
\caption{Sequence diagram of the SL smart contract and token economy smart contract, with an orchestrator and exemplary two institutions (I1 and I2) interacting.}
\end{figure}

For the integration of the data point valuation into swarm learning, we focus on a blockchain-based implementation of swarm learning. Thereby, we develop and deploy two smart contracts: one for the swarm learning and a subsequent data valuation through DDVal on a private blockchain, and one for the token economy on a public blockchain. Clients interact with both and thereby keep states in both synchronized. We illustrate the sequence in Figure~\ref{fig:sequence}, and describe the sequence in the following.

After the orchestrator deploys smart contracts on both blockchains, the private and public one, the orchestrator directly submits the contract addresses to all participating institution servers. These then join the smart contracts. The orchestrator then funds the public smart contract, and afterward, the institutions conduct the SL and data valuation, communicating via the private blockchain and IPFS. In this process, we measure the passed time, which is primarily driven by synchronization wait times (i.e., waiting for all institutions to upload new gradient hashes), and file transfer times. After finishing the SL and data valuation, the institutions publish the institutional contribution quantifications on the public blockchain. Once all are published, the funds are distributed.

Through the use of a private blockchain and IPFS network, only authorized institutions can observe the learning activity in terms of metadata (i.e., gradient hashes, learning rounds), and ML model data. On the public blockchain, the public can observe the test results and fund flows. This can improve the trust of individuals in cases where institutions promise to pass their revenues onward to data providers, like hospitals to patients. At the same time, it leaves more sensitive data like ML models itself shielded from the public. The public smart contract itself comprises 83 lines of code, as such, it is auditable which can increase the trust of entities in it.

When implemented, the transaction costs for transactions on the public Ethereum blockchain are measured in terms of gas cost. Thereby, we split the transactions regarding the user roles. The orchestrator creates the smart contract in the first place, then funds it, and then initiates the payout. The institution only reports the contribution indices. By far the highest gas consumption was for the contract creation, with a relatively large value of 1,473,028. At the time of writing on April 7, 2023, the gas price on the public Ethereum mainnet is $33$ Gwei, where $1$ Gwei equals $10^{-9}$ Ether, and the Ether price is $1859$ USD. Ether is the main cryptocurrency and a core component of the Ethereum blockchain. In total, the cost to deploy the token economy smart contract is ca. $90.37$ USD on the public Ethereum blockchain for 3 institutions. Adding additional institutions scales linearly with 34,399 gas consumption or ca. $1.50$ USD per additional institution. The other transaction costs are much lower, for the orchestrator in this case $3.20$ USD for the contract funding, $7.24$ USD for the contract payout. For each of the institutions, the cost to post the SVs to the smart contract is $7.17$ USD.

Thereby, the process with communicating models, gradients and deep features was relatively fast and completed within ca. 5 minutes. This is much faster than the processes of training the ML models itself, which can take many hours. Thus, the network overhead can be considered as small.

\section{Discussion}\label{sec:discussion}

\subsection{Privacy aspects}\label{section:privacy}
The core idea of using FL and SL is the enabling of ML across data silos while preserving the information privacy of the data owners. As we extend the concept of SL by a data valuation approach, we discuss the privacy implications in more depth.
Sharing the deep features across clients can potentially increase the risk for membership inference attacks, meaning attacks that determine if a data point (i.e., a medical image) is part of the training process. However, membership inference attacks may not be a significant concern in such a healthcare setting, as two institutions would need the same or a very similar data point of a patient, and the maximum information gained would be that a patient visited both institutions.

More concerning would be reconstruction attacks, in which SL clients could reconstruct training data points (e.g., medical images) from the data. Such reconstruction attacks are difficult here, since only the last output layer of the neural network in an inference with training data points is shared. Contemporary attacks require access to more layers~\cite{balle2022reconstructing}.

Still, there could be ways for additional protection and guarantees against such hypothetically possible attack scenarios. For example, differential privacy (DP) can be implemented, in a similar concept to DP in training ML models~\cite{abadi2016deep}. Thereby, before clients submit the vectors of the last output layer of each training data point to the servers, they can add Gaussian noise to the vectors. The standard deviation of the noise could depend on the standard deviation of each vector dimension, and a scaling factor which controls the level of privacy protection. The mean of the noise is 0. Additionally, a clipping can be implemented. We leave this implementation to future research. An interesting question for future research would be the impact of DP on the SV estimation accuracy, and how it affects SVs (e.g., SVs become more uniform with more noise).

\subsection{Principal findings}

With our research, we develop the concept of data point valuation in decentralized learning and develop a technique named DDVal for it. As such, it is suitable for FL and SL. We empirically show that it works accurately in valuing data by identifying mislabelled data and valuing multidimensional labelled data points on three CXR datasets. Thereby, we find that scans with multiple present medical conditions generally have a higher SV than scans with fewer medical conditions or no medical condition.

We also find that DDVal can help to value contributions hierarchically, for example, on a per person or per institution level. Thereby, the empirical results show that it especially works well also for non-IID data distribution cases and outperforms existing methods for data valuation on an institutional level.

Furthermore, we find that DDVal has interesting scalability properties, as it scales with the number of data points, instead of the number of clients in decentralized learning. As such, it is especially suited for decentralized learning settings with many collaborating clients, and where each client only has limited amounts of data. The loglinear complexity of DDVal can be advantageous over existing approaches with exponential complexity in large FL coalitions.

Additionally, we find that DDVal integrates well into a blockchain-based SL architecture. Thereby, an IPFS network and a smart contract on a private blockchain can not only control the SL process, but also the subsequent data valuation. Furthermore, another smart contract can control the token economy and pay out funds to institutions proportionally to their contributions. We find running the data valuation in a decentralized manner comes with a reasonable time-overhead and reasonable cost for running the smart contract on the public blockchain, showing the practicality of the approach.

\subsection{Implications for research and practice}

Our research provides manifold implications for research and practice. For research, we develop the concept of data point valuation in FL and SL and show useful applications in rewarding individuals for providing data to ML tasks and identifying mislabelled data points. As such, data point valuation in decentralized ML has the potential to become a more active field of research. Thereby, we demonstrate that a KNN SV approximation method can successfully value data points, and it can also help to draw conclusions about the contribution values of institutions. Thus, we enable new ways for data valuation in collaborative ML, which has traditionally relied on techniques such as pooling of data in trusted execution environments~\cite{hynes2018demonstration}.
Furthermore, we demonstrate promising applications of data point valuation for FL in healthcare, such as rewarding institutions and individuals for providing their data and detecting likely mislabelled data instances. Moreover, we raise awareness on the importance of having data valuation techniques that can also cope with non-IID data distributions in decentralized learning.
Furthermore, we show that research on decentralized system architectures can not only incorporate the collaborative learning process, but also a subsequent data valuation and reward distribution. Thereby, machine learning aspects and distributed system aspects need to be covered holistically to yield optimal system designs.

For practice, our results show that novel data valuation techniques are ready for various applications, such as rewarding of patients and institutions in healthcare for providing data, detecting good data sources, and identifying mislabelled data. Furthermore, we showcase that SL extended by a data valuation and reward systems can help to enable the vision of web3 in practice to reward users of information systems for providing data.

\subsection{Limitations and future research}
Our research has several limitations that enable possibilities for future research. First, while we use three different datasets for the evaluation, the three datasets are all CXR datasets. Thus, future research could focus on more medical and non-medical datasets for evaluation. Second, DDVal relies on sharing more information than canonical FL and SL approaches, which could have privacy implications. We discussed this in Section~\ref{section:privacy}. While we believe the privacy implications are fine for certain use cases, future research could aim to either improve upon the privacy, or provide privacy guarantees, through techniques such as differential privacy or secure multiparty computation. For differential privacy, it will be interesting to observe if it reduces the accuracy of the SV estimations, and if the SVs equalize.
Third, it remains open how organizations and users perceive DDVal and what the actual practical benefits are. Thus, future research could aim to apply our DDVal in practice and study the real-world adoption of data valuation and related concepts.

\section{Conclusion}\label{sec:conclusion}

In this research, we developed a novel approach capable of valuing data points in FL and SL based on approximating SVs through sharing deep features named DDVal. Besides valuing data points, DDVal can also help to estimate SV contributions of institutions. In our empirical evaulation, it works equally well for IID and non-IID data distributions with a cosine smiliarity of ca. 99.969 \%. Furthermore, it has desirable scalability properties of a loglinear complexity with regard to the number of data points, and is especially interesting for many clients with few data points each, like in our scalability experiment with 16 clients having 8,000 data points each. We integrate DDVal into a decentralized system, which extends the SL concept by data point valuation and a token economy, and show promising applications. In a healthcare context, for example, DDVal can reward institutions and enable them to reward patients for sharing data to ML tasks. Furthermore, it can help to detect likely mislabelled data. Ultimately, DDVal can help to enable visions of the concept web3 into practice, where users are rewarded for sharing their data with machine learning systems.

\section*{Acknowledgements}
We thank Mathias Lécuyer for the valuable discussions about hierarchical data valuation approaches in FL and privacy aspects. For the computations, the authors acknowledge support by the state of Baden-Württemberg through bwHPC and the German Research Foundation (DFG) through grant INST 35/1597-1 FUGG.

 \bibliographystyle{unsrt}
 \bibliography{main.bib}

\end{document}